\documentclass[conference]{IEEEtran}
\IEEEoverridecommandlockouts

\usepackage{cite}
\usepackage{amsmath,amssymb,amsfonts}
\usepackage{algorithmic}
\usepackage{graphicx}
\usepackage{textcomp}
\usepackage{xcolor}
\usepackage{booktabs}
\def\BibTeX{{\rm B\kern-.05em{\sc i\kern-.025em b}\kern-.08em
    T\kern-.1667em\lower.7ex\hbox{E}\kern-.125emX}}
\begin{document}

\title{SpectraNet: FFT-assisted Deep Learning Classifier for Deepfake Face Detection\\
\thanks{}
}

\author{
    \IEEEauthorblockN{
        Nithira Jayarathne\textsuperscript{1},
        Naveen Basnayake\textsuperscript{1},
        Keshawa Jayasundara\textsuperscript{1},
        Pasindu Dodampegama\textsuperscript{1},\\
        Praveen Wijesinghe\textsuperscript{1},
        Hirushika Pelagewatta\textsuperscript{1},
        Kavishka Abeywardana\textsuperscript{1},\\
        Sandushan Ranaweera\textsuperscript{2},
        Chamira Edussooriya\textsuperscript{1}
    }
    \\
    \IEEEauthorblockA{
        \textsuperscript{1}Department of Electronic and Telecommunications Engineering, University of Moratuwa\\
        \textsuperscript{2} School of Electrical and Data Engineering, University of Technology Sydney\\
    }
}

\maketitle

\begin{abstract}
Detecting deepfake images is crucial in combating misinformation. We present a lightweight, generalizable binary classification model based on EfficientNet-B6, fine-tuned with transformation techniques to address severe class imbalances. By leveraging robust preprocessing, oversampling, and optimization strategies, our model achieves high accuracy, stability, and generalization. While incorporating Fourier transform-based phase and amplitude features showed minimal impact, our proposed framework helps non-experts to effectively identify deepfake images, making significant strides toward accessible and reliable deepfake detection.
\end{abstract}

\section{Introduction}

Advances in synthetic data generation have introduced both opportunities and challenges, particularly with the rise of deepfake technology. While useful in creative industries, deepfakes present significant threats such as public opinion manipulation, identity theft, and geopolitical instability. Effective detection methods are essential to mitigate these risks.\cite{https://doi.org/10.13140/rg.2.2.23089.81762}

Facial deepfakes are especially prone to misuse and can be categorized into three types: face reenactment, which alters a target image using source facial features; face swapping, which places known individuals in fabricated contexts; and face synthesis, which generates entirely artificial faces\cite{Deepfake_Generation_and_Detection:_A_Benchmark_and_Survey}. The accessibility of tools for creating such images has heightened the need for robust detection solutions.

We propose a novel approach to detecting facial deepfake images using an EfficientNet-B6 \cite {EfficientNet:_Rethinking_Model_Scaling_for_Convolutional_Neural_Networks} model and a hybrid method combining Fourier Transform \cite{romeo2024fasterliesrealtimedeepfake} with EfficientNet-B6. The Fourier Transform leverages phase and amplitude analysis to detect inherent smoothness in synthetic images. However, our experiments reveal that while the Fourier Transform adds a theoretical advantage, its practical impact on accuracy is limited. Our EfficientNet-B6 model, optimized with additional hyperparameter tuning, achieved a significant accuracy of 91\%. To address class imbalance in the dataset, we employed image transformation techniques, effectively increasing the representation of true images and enhancing the training process.

The proposed model distinguishes itself from existing methods by its lightweight architecture, relatively high accuracy, and improved generalizability across different datasets. We used a comprehensive evaluation framework including metrics such as AUC, accuracy, F1-score, precision, and recall to validate the performance of the models. This research highlights the potential of EfficientNet-B6 in deepfake detection while providing insights into the limited effectiveness of Fourier Transform in this context. Our findings contribute to the growing body of work on lightweight and robust deepfake detection systems, paving the way for practical and scalable solutions in real-world applications.

\begin{figure}[t]
    \centering
    \includegraphics[width=\linewidth]{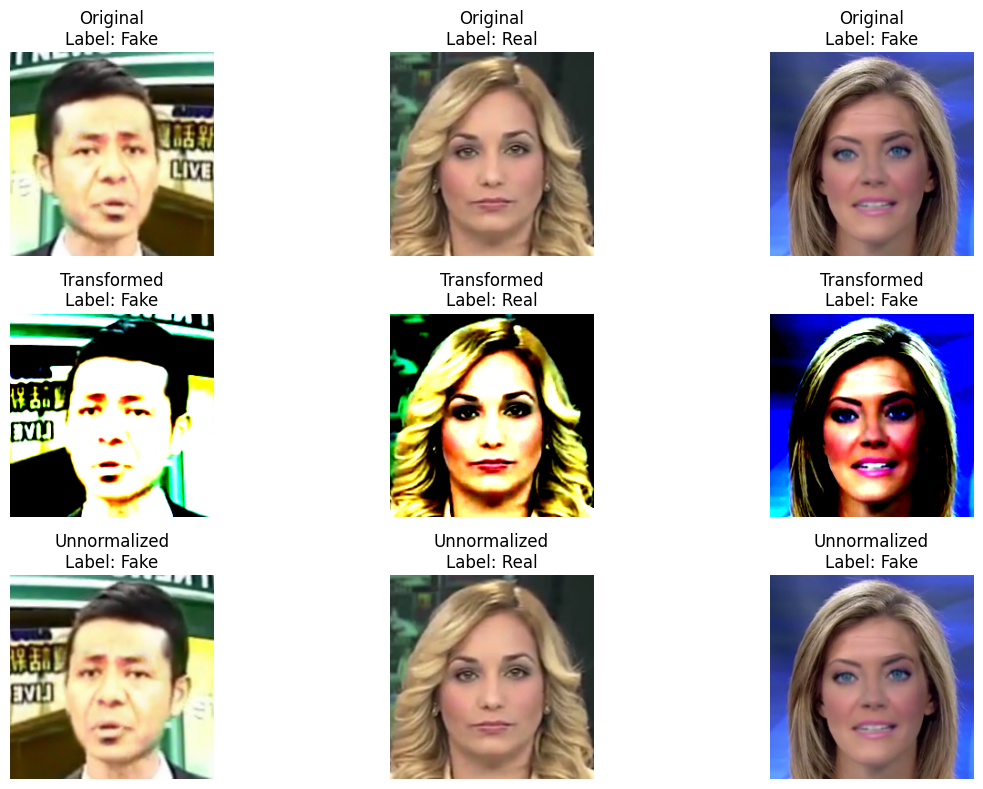} 
    \caption{Transformed instances of input images, including augmentations and preprocessing steps}
    \label{fig:example_image}
\end{figure}

\section{Methodology}

The proposed methodology involves dataset preparation, adaptation of the EfficientNet-B6 architecture, and the development of a strong training framework for binary classification. EfficientNet-B6 was selected for its lightweight design and fast inference time. To address the class imbalance in the dataset, oversampling techniques were applied to ensure an equal distribution of real and fake images. The model, known for its efficient feature extraction and compound scaling, was fine-tuned for the task. Advanced optimization techniques were used to improve accuracy, stability, and generalization. 
\\The Adam optimizer (Adaptive Moment Estimation)\cite{kingma2017adammethodstochasticoptimization} was chosen for its stability and efficient handling of sparse gradients. The ReduceLROnPlateau scheduler was used to adjust the learning rate, promoting faster convergence and reducing overfitting. Binary Cross-Entropy with Logits Loss (BCEWithLogitsLoss) was selected as the loss function, combining binary cross-entropy and sigmoid activation for effective binary classification.

\subsection{Data Preparation and Preprocessing}
The dataset comprises a total of 262,160 images, with 42,690 labeled as real and 219,470 labeled as fake, resulting in a significant class imbalance. To address this imbalance, In each epoch we use 25600 images, with a batch size of 256 where half of the images are taken randomly from the real data set and the other half is taken randomly the fake data set. We repeat this process for each epoch.\cite{DeepfakeBench_YAN_NEURIPS2023}\cite{Celeb-Df}\cite{FaceForensics++}\cite{DFDC}

This batching approach ensures efficient utilization of GPU memory and facilitates smoother training.
By creating balanced batches and leveraging the newly augmented dataset, we improve the model's exposure to real and fake image samples during training, mitigating the effects of class imbalance and enhancing overall training efficiency

During the training loop, the images undergo the following preprocessing steps:
\begin{enumerate}
    \item \textbf{Resizing:} Images are resized to a consistent dimension to standardize input data and optimize computational efficiency.
    \item \textbf{Data Augmentation:} Techniques such as horizontal flipping are applied to artificially expand the dataset, introducing variability that reduces overfitting.
    \item \textbf{Normalization:} Images are converted to tensors and normalized to a standard range, facilitating stable gradient descent by reducing numerical variability in feature scales.
\end{enumerate}

This process is repeated for each epoch, ensuring that the model is exposed to a diverse and balanced dataset during training, thereby improving generalization.

\subsection{Model Architecture and Adaptation}
We employ EfficientNet-B6  \cite {EfficientNet:_Rethinking_Model_Scaling_for_Convolutional_Neural_Networks}, a convolutional neural network architecture that utilizes a compound scaling methodology to balance network depth, width, and resolution, optimizing accuracy while maintaining computational efficiency. Pretrained on the ImageNet dataset, EfficientNet-B6 provides robust feature extraction capabilities derived from over 14 million labeled images.

For the binary classification task, the model is adapted as follows:
\begin{enumerate}
    \item \textbf{Fine-Tuning the Pretrained Weights:}
    \begin{itemize}
        \item \textbf{Unfreezing All Layers:} All layers of the pretrained model are unfrozen, enabling fine-tuning of both shallow and deep layers. This ensures that feature representations are adjusted to capture domain-specific patterns in the new dataset.
        \item \textbf{Modification of the Final Layers:}
        \begin{itemize}
            \item A dropout layer is added to mitigate overfitting by stochastically deactivating neurons during training.
            \item A dense linear layer configured to output a single logit is introduced, enabling binary classification.
        \end{itemize}
        \item \textbf{Integration of Pretrained Knowledge:} By starting with ImageNet-pretrained weights, the model benefits from robust feature extraction capabilities. These pretrained weights provide a strong initialization for extracting both low-level features (e.g., edges and textures) and high-level semantic representations (e.g., object parts).
    \end{itemize}
\end{enumerate}

This fine-tuning approach ensures the EfficientNet-B6 model effectively balances the generalization capacity of its pretrained backbone with the specificity required for the binary classification task.

\subsection{Training Framework and Optimization}
To ensure a robust and efficient training process, the following methodologies and tools are employed:
\begin{enumerate}
    \item \textbf{Loss Function:} Binary Cross-Entropy with Logits Loss (BCEWithLogitsLoss) is used for classification. This loss function combines the sigmoid activation function and cross-entropy loss in a single computation, enhancing numerical stability by reducing potential gradient saturation or vanishing issues during training.
    \item \textbf{Optimizer:} The Adam optimizer is employed due to its adaptive learning rate and momentum-based parameter updates, offering a balance between convergence speed and optimization stability. L2 regularization (weight decay) is applied to penalize excessively large weights, thereby reducing the risk of overfitting.
    \item \textbf{Learning Rate Scheduler:} The ReduceLROnPlateau scheduler dynamically adjusts the learning rate based on the validation loss. The initial learning rate is set to \(5 \times 10^{-4}\). When the validation loss stagnates or shows no significant improvement for a predefined number of epochs, the scheduler reduces the learning rate by a factor, allowing steady convergence without abrupt oscillations in learning rates.
    \item \textbf{Mixed Precision Training:} Mixed precision training \cite{micikevicius2018mixedprecisiontraining}optimizes computational efficiency and reduces memory consumption by employing 16-bit floating-point arithmetic for most operations while maintaining 32-bit precision for critical updates, such as gradient scaling.
    \begin{itemize}
        \item The PyTorch AMP (Automatic Mixed Precision) toolkit’s GradScaler function dynamically scales gradients, preventing numerical underflow or overflow during computation.
        \item This approach accelerates training on modern GPUs while preserving the stability and precision required for convergence.
    \end{itemize}
\end{enumerate}

By combining rigorous data preparation, a carefully adapted EfficientNet-B6 architecture, and a robust training framework, we successfully addressed the challenges posed by class imbalance and domain-specific requirements in the binary classification task. The integration of oversampling techniques, fine-tuning of pretrained weights, and advanced optimization strategies ensures that the model is both computationally efficient and capable of achieving high accuracy. This comprehensive approach highlights the synergy between modern deep learning architectures and innovative training methodologies in tackling complex classification problems.

\section{Experiments and Results}

The validation set provided was evaluated, and the following evaluation metrics were calculated.

\subsection{Selection of Metrics for Evaluation}
To evaluate the performance of our model comprehensively, we have chosen a set of robust evaluation metrics. These metrics assess the model's effectiveness in distinguishing between real and fake images and provide insights into various aspects of its classification performance.\cite{powers2020evaluationprecisionrecallfmeasure} Below is a brief description of each metric:

\begin{itemize}
    \item \textbf{Area Under the Curve (AUC):} The AUC metric measures the area under the Receiver Operating Characteristic (ROC) curve, which plots the true positive rate (TPR) against the false positive rate (FPR) at various classification thresholds. AUC provides a single scalar value summarizing the model's ability to distinguish between classes.
    \begin{equation}
        \text{AUC} = \int_0^1 \text{TPR}(\text{FPR}) \, d(\text{FPR}),
    \end{equation}
    where 
    \[
        \text{TPR} = \frac{\text{TP}}{\text{TP} + \text{FN}}, \quad \text{FPR} = \frac{\text{FP}}{\text{FP} + \text{TN}}.
    \]
    A higher AUC indicates better discrimination ability, with 1 representing perfect classification and 0.5 indicating random guessing.

    \item \textbf{Accuracy:} Accuracy quantifies the proportion of correctly classified images among all images in the dataset. It is calculated as:
    \begin{equation}
        \text{Accuracy} = \frac{\text{True Positives} + \text{True Negatives}}{\text{Total Number of Samples}}.
    \end{equation}
    While accuracy is intuitive and widely used, it can be misleading for imbalanced datasets, where other metrics such as AUC, precision, and recall provide additional context.

    \item \textbf{F1 Score:} The F1 Score is the harmonic mean of precision and recall, providing a balanced measure that accounts for both false positives and false negatives. It is calculated as:
    \begin{equation}
        \text{F1} = 2 \cdot \frac{\text{Precision} \cdot \text{Recall}}{\text{Precision} + \text{Recall}}.
    \end{equation}

    \item \textbf{Precision:} Precision measures the proportion of correctly classified positive samples (real images) out of all samples predicted as positive. It is calculated as:
    \begin{equation}
        \text{Precision} = \frac{\text{True Positives}}{\text{True Positives} + \text{False Positives}}.
    \end{equation}

    \item \textbf{Recall (Sensitivity):} Recall, also known as sensitivity, measures the proportion of correctly identified positive samples out of all actual positive samples. It is calculated as:
    \begin{equation}
        \text{Recall} = \frac{\text{True Positives}}{\text{True Positives} + \text{False Negatives}}.
    \end{equation}
\end{itemize}

By using this diverse set of metrics, we aim to provide a comprehensive evaluation of our model's performance, addressing both its strengths and potential weaknesses in handling the classification task. This approach ensures that the evaluation captures various aspects of model behavior, offering a nuanced understanding of its capabilities.

\subsection{Classification of Images}
For the classification task, we utilized two deep learning models:
\begin{enumerate}
    \item \textbf{EfficientNet-B6 Model (Our Proposed Method):} In this approach, the EfficientNet-B6 model was fine-tuned on the new dataset. All layers of the model were trained, with the final linear layer (originally designed for the 1,000-class ImageNet classification task) replaced by a dropout layer to reduce overfitting and a new linear layer tailored for binary classification.

\begin{figure}[h]
    \centering
    \includegraphics[width=1.1\linewidth]{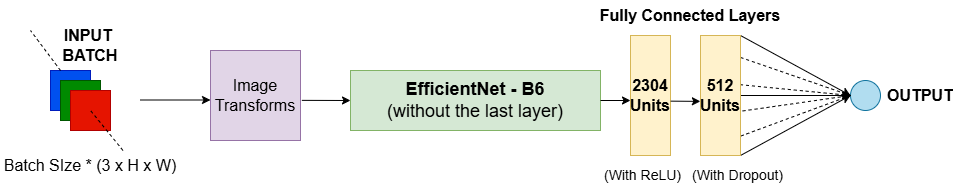} 
    \caption{EfficientNet-B6 model architecture for direct classification of deepfake images.}
    \label{fig:example_image}
\end{figure}

    \item \textbf{Hybrid Model:} This model combines the EfficientNet-B6 architecture with a frequency model \cite{Using_fourier_transform} to enhance classification performance. The EfficientNet-B6 was trained on the dataset to extract spatial features, while the frequency model was used to analyze frequency-domain information, such as phase and amplitude. The outputs from the EfficientNet-B6 and the frequency model were concatenated to provide a richer feature representation.
    
\begin{figure}[h]
    \centering
    \includegraphics[width=1.1\linewidth]{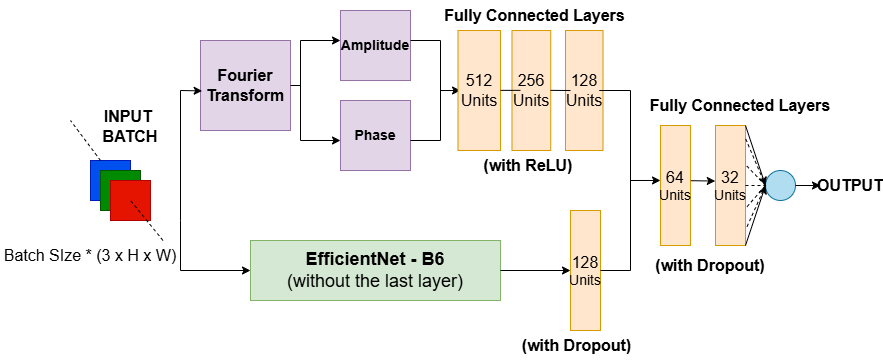} 
    \caption{Hybrid model integrating EfficientNet-B6 with Fourier Transform phase and amplitude information}
    \label{fig:example_image}
\end{figure}

\end{enumerate}

\subsection{Results and Analysis}
The following table presents the classification results obtained from the two models:

\begin{table}[]
\centering
\caption{Performance metrics for EfficientNet-B6 and the Hybrid Model. \textit{P}-\textit{precision}, \textit{R}-\textit{recall}, \textit{AUC}-\textit{Area Under Curve}, \textit{Acc}-\textit{accuracy}.}
\begin{tabular}{lccccc}
\toprule
\textbf{Method} & \textbf{AUC} & \textbf{ACC} & \textbf{F1} & \textbf{P} & \textbf{R} \\
\midrule
Hybrid model & 0.8984 & 0.8981 & 0.8946 & 0.9346 & 0.8579 \\
\textbf{EfficientNet-B6} & \textbf{0.9104} & \textbf{0.9102} & \textbf{0.9074} & \textbf{0.9435} & \textbf{0.8740} \\
\bottomrule
\end{tabular}

\label{tab:results}
\end{table}
As presented in Table I, the EfficientNet-B6 model achieved impressive results, attaining an accuracy and AUC (Area Under the Curve) score of 0.9102. However, the integration of Fourier transform phase and amplitude information led to underperformance across all evaluated metrics. This outcome suggests that deepfake generation techniques effectively encode and replicate the frequency components of facial images, limiting the utility of frequency-based information as a distinguishing factor. In this context, EfficientNet-B6 has demonstrated its strength as a robust feature extractor for image-based classification tasks. Among the tested models, including EfficientNet-B0, B3, and B6, the B6 variant consistently yielded the best performance, underscoring its suitability for this application.
\begin{table}[]
\centering
\caption{Evaluation Time for 3072 Test Files}
\begin{tabular}{lccccc}
\toprule
\textbf{Method} & \textbf{Time (seconds)}\\
\midrule
Hybrid model & 3.48  \\
\textbf{EfficientNet-B6} & \textbf{2.55} \\
\bottomrule
\end{tabular}

\label{tab:results}
\end{table}

According to Table II, the EfficientNet-B6 model demonstrates a significantly lower evaluation time of 2.55 seconds. In contrast, the Hybrid model requires considerably more time at 3.48 seconds, attributed to its greater model complexity. The evaluations were conducted using an NVIDIA RTX 4080 GPU. The shorter evaluation time of EfficientNet-B6 enhances its suitability for real-world applications where speed and efficiency are critical.

\section{Conclusion}
In this work, we successfully addressed the challenge of class imbalance in a deepfake detection task by utilizing a robust dataset augmentation strategy and adapting the EfficientNet-B6 model. By fine-tuning the pretrained model and employing advanced optimization techniques, we achieved efficient training and high accuracy in distinguishing between real and fake images. The integration of oversampling, mixed precision training, and careful preprocessing ensured improved model generalization. Our evaluation metrics, including AUC, accuracy, precision, recall, and F1 score, demonstrated the model's strong performance. The results highlight the effectiveness of combining modern deep learning architectures with innovative training strategies to solve complex classification problems.

\bibliographystyle{IEEEtran}
\bibliography{ref1} 

\end{document}